\documentclass[10pt,twocolumn,letterpaper]{article}

\PassOptionsToPackage{table}{xcolor}
\usepackage{cvpr}
\usepackage{graphicx}
\usepackage{amsmath}
\usepackage{amssymb}

\usepackage{multirow}
\usepackage{makecell}
\usepackage[pagebackref,breaklinks,colorlinks]{hyperref}

\title{Face Identity Unlearning for Retrieval via Embedding Dispersion}

\author{
Mikhail Zakharov\thanks{Corresponding author} \\
Kryptonite AI Laboratory, Kryptonite JSC, Moscow, Russia \\
{\tt\small m.zakharov@kryptonite.ru}
}

\begin{document}

\maketitle

\begin{abstract}
Face recognition systems rely on learning highly discriminative and compact identity clusters to enable accurate retrieval. However, as with other surveillance-oriented technologies, such systems raise serious privacy concerns due to their potential for unauthorized identity tracking. While several works have explored machine unlearning as a means of privacy protection, their applicability to face retrieval — especially for modern embedding-based recognition models — remains largely unexplored. In this work, we study the problem of face identity unlearning for retrieval systems and present its inherent challenges. The goal is to make selected identities unretrievable by dispersing their embeddings on the hypersphere and preventing the formation of compact identity clusters that enable re-identification in the gallery. The primary challenge is to achieve this forgetting effect while preserving the discriminative structure of the embedding space and the retrieval performance of the model for the remaining identities. To address this, we evaluate several existing approximate class unlearning methods (e.g., Random Labeling, Gradient Ascent, Boundary Unlearning, and other recent approaches) in the context of face retrieval and propose a simple yet effective dispersion-based unlearning approach. Extensive experiments on standard benchmarks (VGGFace2, CelebA) demonstrate that our method achieves superior forgetting behavior while preserving retrieval utility.
\end{abstract}

\section{Introduction}
Face recognition and retrieval systems have become essential components in modern surveillance, authentication and security technologies. These systems rely on deep embedding models~\cite{taigman2014deepface, schroff2015facenet, Deng2018} that map aligned facial images to highly discriminative and compact representations, enabling efficient retrieval and robust recognition across different large-scale galleries. However, this ability to maintain consistent identity in embeddings across different views raises significant privacy concerns, as it allows persistent and potentially unauthorized identity tracking, leading to ethical and societal implications.

Recent efforts in machine unlearning~\cite{cao2015unlearning, ginart2019making, sekhari2021remember} aim to address privacy concerns by enabling models to selectively forget specific data, aligning with the \textit{“right to be forgotten”}~\cite{Hoofnagle02012019} principles in modern privacy regulations. In this context, the General Data Protection Regulation (GDPR) compliance involves not only removing the data itself but also eliminating its influence on the model’s parameters. However, our objective differs: rather than merely erasing the data’s contribution, we seek to make the representations of the forgotten identities non-discriminative — effectively disrupting their compact cluster structure on the identity manifold. While a variety of unlearning algorithms have been explored in classification and image generation tasks~\cite{fan2024salun, graves2021amnesiac, kurmanji2023towards}, their applicability to face retrieval has been largely unexplored. Unlike classification, where forgetting can be expressed as removing or shrinking a class decision boundary, retrieval requires maintaining a globally structured embedding space, making selective forgetting a far more delicate and challenging task.

In this work, we introduce and systematically study the problem of face identity unlearning for deep face embedding models. The objective is to make selected identities unretrievable — essentially dispersing their embeddings on the hypersphere to prevent the formation of compact identity clusters. The goal of this dispersing process is to achieve that facial images of the same identity, captured under different views, are no longer mapped to nearby locations on the hypersphere in terms of angular distance. The key challenge is to achieve this forgetting effect while preserving the discriminative structure of the overall embedding space and maintaining high retrieval performance for non-forgotten identities.

We revisit several \textit{class unlearning algorithms} — including Random Labeling~\cite{golatkar2020eternal, hayase2020selective}, Gradient Ascent~\cite{golatkar2020eternal}, Boundary Unlearning~\cite{chen2023boundary}, along with several recent \textit{approximate unlearning approaches}~\cite{foster2024zeroshot, kravets2025zeroshot, chopra2005learning} — and adapt these methods to deep face embedding models by reformulating them for face-specific loss functions such as CosFace~\cite{Wang2018}. We evaluate their performance using standard retrieval metrics (mAP and CMC-Rank-1) as well as a newly introduced cluster compactness measure. Our findings reveal that while these methods can reduce classification accuracy for the forgotten identities even in challenging face recognition scenarios, they fail to adequately disperse their embeddings, leaving identity clusters largely intact and retrievable.

Motivated by these findings, we propose a simple yet effective \textbf{dispersion-based unlearning algorithm}. This method explicitly minimizes the cosine similarities between embeddings within the same compact identity cluster, thereby breaking up its compact structure while leaving the remaining clusters and the overall embedding space largely unaffected. 

Extensive experiments on multiple benchmarks — including VGGFace2~\cite{cao2018vggface2}, CelebA~\cite{liu2015faceattributes}, CFP~\cite{cfp-paper}, and a subset of IJB-B~\cite{ijbb2017} — demonstrate that, despite its simplicity, our approach achieves a significantly higher degree of forgetting — measured via retrieval and classification metrics, as well as embedding visualizations — while maintaining competitive performance on the retained identities.

\textbf{We summarize our major contributions as follows:}
\begin{itemize}
    \item We introduce and formalize the problem of face identity unlearning for deep face embedding models, highlighting its unique challenges compared to classification-based unlearning.
    \item We provide a comprehensive evaluation of existing approximate unlearning methods in this context, revealing their limitations in disrupting the embedding-level identity structure.
    \item We propose a simple yet effective dispersion-based unlearning approach that explicitly aims to destroy compact identity clusters, achieving superior forgetting while preserving retrieval accuracy for the remaining identities.
\end{itemize}

\section{Related Work}
\textbf{Face Recognition.} Face recognition is one of the most extensively studied topics in modern deep learning, addressing the problem of open-set recognition — recognizing and distinguishing identities unseen during training. The dominant approach relies on deep embedding-based models that learn highly discriminative facial representations, enabling efficient identification, verification, and retrieval through similarity search in a high-dimensional embedding space.

Early breakthroughs were achieved by deep convolutional models such as DeepFace~\cite{taigman2014deepface}, DeepID~\cite{sun2015deepid3}, and FaceNet~\cite{schroff2015facenet}, which demonstrated remarkable performance across various face-related tasks. Beyond the expansion of large-scale datasets (e.g., VGGFace2~\cite{cao2018vggface2}, MS-Celeb-1M~\cite{guo2016msceleb}, CelebA~\cite{liu2015faceattributes}) and architectural refinements~\cite{simonyan2015vgg, szegedy2015inception, he2016resnet, dosovitskiy2021vit, zhao2021facetransformer}, much of the recent progress in face recognition has been driven by advances in loss design that enforce stronger intra-class compactness and inter-class separability. Initial efforts employed metric-learning–based losses, including the contrastive loss~\cite{chopra2005learning}, triplet loss~\cite{schroff2015facenet}, and angular loss~\cite{wang2017deep}. However, due to the combinatorial explosion of possible sample triplets, these methods proved computationally inefficient for large-scale datasets. Subsequent research therefore focused on more scalable and efficient classification-based losses.

Wen \textit{et al.}~\cite{wen2016discriminative} introduced the center loss to improve intra-class compactness and inter-class separability by learning a representative center for each identity. The \( L_{2} \)-softmax~\cite{Ranjan2017} explored the effects of normalization on both embeddings and classifier weights. Later, several angular margin-based losses — including SphereFace~\cite{Liu2017}, AM-Softmax~\cite{wang2018additive}, CosFace~\cite{Wang2018}, and ArcFace~\cite{Deng2018} — achieved significant improvements in generalization and scalability, effectively defining the modern paradigm of deep face recognition. Building upon these, several variants have been proposed, such as MagFace\cite{meng2021magface}, AdaFace~\cite{kim2022adaface}, and CurricularFace~\cite{huang2020curricularface}, introducing magnitude-aware feature regularization, adaptive margin adjustment, and curriculum-based training schedules, respectively. Some studies have also explored the integration of Vision Transformers (ViT)~\cite{dosovitskiy2021vit} as backbones for face recognition, e.g., Face-Transformer~\cite{zhao2021facetransformer} and related works.

In our work, we adopt a CNN-based embedding model trained with CosFace~\cite{Wang2018}, which remains a widely used industry standard in popular open-source benchmarks such as \textit{InsightFace}~\cite{insightface2019}. We note that substituting the angular margin-based loss (e.g., ArcFace~\cite{Deng2018} or AM-Softmax~\cite{wang2018additive}) would not qualitatively affect our findings, as our analysis focuses on the unlearning dynamics of compact identity clusters rather than the specific loss formulation.

\smallskip
\textbf{Machine Unlearning.} Machine unlearning aims to remove the influence of specific data samples from a trained model such that the model behaves as if it has never been exposed to this data during training. This concept enables users to erase their personal information from learned models and was originally motivated by data protection regulations such as the \textit{“right to be forgotten”} principle~\cite{Hoofnagle02012019}. The term machine unlearning was first introduced by Cao and Yang~\cite{cao2015unlearning}. A probabilistic definition of unlearning was later proposed by Ginart \textit{et al.}~\cite{ginart2019making} and subsequently extended in several works, including Sekhari \textit{et al.}~\cite{sekhari2021remember}, Gupta \textit{et al.}~\cite{gupta2021adaptive}, and Neel \textit{et al.}~\cite{neel2021descent}, among others.

Existing machine unlearning methods can be broadly divided into two categories: \textit{exact} and \textit{approximate} unlearning. Exact unlearning focuses on developing algorithms with theoretical guarantees or formal unlearning certificates. Representative approaches include differential privacy (DP)-enforced unlearning and certified data removal methods~\cite{chien2022certified, guo2020certified}. Another widely known approach, SISA~\cite{bourtoule2021machine} , partitions the dataset into multiple shards and trains a separate model for each subset. When an unlearning request is received, only the affected models are retrained. \textit{Retraining} a model from scratch after removing the forget set from training data is considered the gold standard for exact unlearning~\cite{thudi2022necessity}, but such methods are computationally prohibitive and generally impractical for large-scale deep models.

Approximate unlearning has emerged as a more practical alternative, aiming for fast and effective removal of information without theoretical guarantees. Instead, empirical evaluations are used to assess privacy guarantees, often via such metrics as membership inference attacks (MIA)~\cite{carlini2021membership}. Golatkar \textit{et al.}~\cite{golatkar2020eternal} introduced Fisher Forgetting, which removes the influence of forgetting data by applying a noisy Newton update on model parameters, where the noise is derived from the Fisher Information Matrix (FIM) to target data-specific knowledge. SCRUB~\cite{kurmanji2023towards} adopts a teacher–student framework in which the teacher transfers knowledge to the student model while excluding the forget-set samples or classes. Amnesiac Unlearning~\cite{graves2021amnesiac} employs randomized label training for the forget set, followed by fine-tuning on the retain set to repair the model’s utility. Random Labeling~\cite{golatkar2020eternal, hayase2020selective} and Gradient Ascent~\cite{golatkar2020eternal} perform fine-tuning on the forget samples using random labels or gradient ascent directions, respectively. Boundary unlearning~\cite{chen2023boundary} unlearns an entire class by shifting decision boundaries. More recent studies explore Lipschitz-regularized unlearning~\cite{foster2024zeroshot, kravets2025zeroshot}, which encourages smooth model responses to forget-set perturbations by injecting Gaussian noise of the same dimensionality into the inputs and minimizing the ratio between output and input changes. Contrastive Unlearning~\cite{lee2025contrastive} formulates unlearning directly in the representation space by contrasting embeddings of forget samples against those of retained samples, leveraging the geometric properties of the embedding space. Finally, SalUn~\cite{fan2024salun} introduces a “plug-and-play” enhancement for existing methods by leveraging weight saliency, updating only parameters considered important for effective unlearning.

Our work is different from existing machine unlearning approaches. Even if the model never observes the forgotten identities during training, a deep embedding-based model can still successfully retrieve and match the same identity across different views, as their embeddings naturally form compact clusters in the latent space. Consequently, most existing unlearning methods are not applicable in this setting. While approaches that maximize the errors on the forget set (e.g., ~\cite{golatkar2020eternal, chen2023boundary}) can be applied, our findings show that, being primarily designed for classification tasks, they fail to effectively break the compact cluster structure of forgotten identities and thus remain insufficient for disrupting retrieval performance.

\section{Preliminaries and Problem Formulation}
\subsection{Face Embedding Model}
We consider a face embedding model $f_{\omega}$, a deep neural network-based encoder parametrized by weights $\omega$, that maps an input facial image $I_i$ to a $d$-dimensional embedding vector $x_i$:
\begin{equation}
x_i = f_\omega(I_i), \quad 
x_i \in \mathbb{R}^{d}, \quad 
I_i \in \mathbb{R}^{H\times W\times 3}.
\label{eq:encoder}
\end{equation}
Here, $I_i$ denotes an aligned facial image of size $H \times W$. The model $f_{\omega}$ is trained 
via backpropagation on large-scale face recognition datasets (e.g., Glint360K~\cite{an2021partial}, VGGFace2~\cite{cao2018vggface2}), using the CosFace loss~\cite{Wang2018}:
\begin{equation}
\mathcal{L}_{\text{CosFace}} = 
-\log
\frac{e^{s(\cos\theta_{y_i} - m)}}
{e^{s(\cos\theta_{y_i} - m)} + \sum_{j \neq y_i} e^{s\cos\theta_j}},
\label{eq:cosface}
\end{equation}
subject to
\begin{equation}
\cos\theta_j = \hat{W}_j^\top \hat{x}_i, \quad
\hat{x}_i = \frac{x_i}{\|x_i\|}, \quad
\hat{W}_j = \frac{W_j}{\|W_j\|}.
\label{eq:cosine}
\end{equation}
where $x_i$ is the $i$-th embedding corresponding to the ground-truth class of $y_i$, $W_j$ is the weight vector of the $j$-th class, $\theta_j$ is the angle between $W_j$ and $x_i$, $s$ is a scaling factor and $m$ is an angular margin. This margin-based formulation encourages compact identity clusters while maximizing the angular separation between different identities in the embedding space.

\subsection{Unlearning Formulation}
We formulate machine unlearning in the context of face identity unlearning using deep neural network–based encoders. Let $\mathcal{D} = \{ (I_i, y_i) \}_{i=1}^N \subseteq \mathcal{X} \times \mathcal{Y}$ denote the training dataset, where $I_i \in \mathcal{X}$ is an aligned facial image and $y_i \in \mathcal{Y}$ is its corresponding identity label. The label space is defined as $\mathcal{Y} = \{1, \ldots, K\}$, where $K$ is the total number of identities in the dataset. In our formulation, forgetting is performed at the \textit{identity level}, i.e., by forgetting entire identity classes rather than individual samples. Let $\mathcal{P}_u \subset \mathcal{Y}$ be the set of identities to be forgotten. Consequently, the forget set is defined as $\mathcal{D}_f = \{ (I_i, y_i) \in \mathcal{D} \mid y_i \in \mathcal{P}_u\}$ and contains facial images of the identities to be forgotten, while the retain set is defined as $\mathcal{D}_r = \{ (I_i, y_i) \in \mathcal{D} \mid y_i \notin \mathcal{P}_u\}$ and contains the remaining data.

Given the face embedding model $f_{\omega}$, our objective is to unlearn the information associated with $\mathcal{D}_f$ by updating its parameters $\omega \rightarrow \omega'$. Formally, this process can be expressed as obtaining $f_{\omega'} = \mathcal{U}(f_{\omega}, \mathcal{D}_f)$, where $\mathcal{U}$ denotes the unlearning algorithm. Following the formulation in~\cite{Peng_2025_CVPR}, a \textit{target unlearning operator} $\mathcal{U}$ should yield a model $f_{\omega'}$ that satisfies the following conditions: 

\begin{equation}
\left\{
\begin{aligned}
\phi_{f_{\omega'}}(I, I_p) &< \phi_{f_{\omega'}}(I, I_n), && 
    \forall I \in \mathcal{P}_u, \; \forall I_n \text{ s.t. } y_n \ne y, \\
\phi_{f_{\omega'}}(I, I_p) &> \phi_{f_{\omega'}}(I, I_n), && 
    \forall I \notin \mathcal{P}_u, \; \forall I_n \text{ s.t. } y_n \ne y.
\end{aligned}
\right.
\end{equation}

Here, $\phi_{f_{\omega'}}(\cdot, \cdot)$ denotes the cosine similarity metric in the embedding space of $f_{\omega'}$, $I$ is an aligned facial image with label $y$, $I_p$ is another facial image of the same identity, and $I_n$ corresponds to  an image of a different identity with label $y_n$. These conditions ensure that the unlearning algorithm $\mathcal{U}$ produces a model $f_{\omega'}$ such that facial images of identities in $\mathcal{P}_u$ become non-discriminative in its representation space, thereby causing an immediate drop in retrieval performance for those identities, while preserving retrieval performance for all remaining ones.

\section{Method}
Our approach utilizes the geometric properties of the embedding space of a well-trained face encoder to perform identity unlearning. Instead of modifying decision boundaries or perturbing model parameters, which effectively serve as proxy objectives for the desired forgetting behavior (i.e., identity unlearning),  we directly target the local geometry of the representation space. Specifically, rather than encouraging embeddings of the same identity to remain close, we aim to disrupt their compactness within the set of forgotten identities $\mathcal{P}_u$, thereby significantly degrading the model’s ability to consistently map images of the same person to neighboring regions. Motivated by general margin-based losses in metric learning, we introduce a simple yet effective \textit{dispersion loss} and its hard-mining variant, the \textit{hard dispersion loss}, which explicitly push same-identity embeddings apart. This approach reduces the model’s capacity to map embeddings of the same identity closely together, effectively dispersing their cluster structure, while leaving the rest of the embedding space largely unaffected.

\subsection{Dispersion Loss}
\label{sec:dispersion}
Given an unlearning batch $\{(I_i, y_i)\}_{i=1}^B$ of size $B$, sampled from the forget set $\mathcal{D}_f$ we compute each embedding $x_i$ using Eq.~\ref{eq:encoder} and obtain its $\ell_2$-normalized version $\hat{x}_i$ as in Eq.~\ref{eq:cosine}. We then disperse these embeddings by minimizing:

\begin{equation}
\mathcal{L}_{\text{disp}} =
\frac{1}{|\mathcal{N}_{\text{pos}}|}
\sum_{\substack{i \in \mathcal{N}_{\text{pos}}}}^{B}
\frac{1}{|P_i|}
\sum_{\substack{j \in P_i}}^{B}
\max\big(0,\, m + \hat{x}_i^\top \hat{x}_j\big),
\label{eq:dispersion}
\end{equation}
where
\[
\begin{aligned}
P_i &= \{\, j \mid y_j = y_i,\, j \neq i \,\},\\
\mathcal{N}_{\text{pos}} &= \{\, i \mid |P_i| > 0 \,\},
\end{aligned}
\]
denote index sets defined within the current batch of size $B$. Specifically, $\mathcal{N}_{\text{pos}}$ contains indices of embeddings that have at least one positive sample in the batch (i.e., another sample of the same identity label $y$), while $P_i$ denotes the set of indices corresponding to positive samples of a given anchor $x_i$ within the batch, and $m$ is the margin. Note that $\hat{x}_i^\top \hat{x}_j$ corresponds to $\cos\theta$, where $\theta$ is the angle between the normalized embeddings. The combination of the parameter $m$ and the operator $max$ defines a lower bound for $\cos\theta$ at $-m$, preventing excessive dispersion that could destabilize the embedding space. Minimizing Eq.~\ref{eq:dispersion} consequently increases $\theta$, effectively disrupting the compactness of identity clusters and degrading retrieval performance on $\mathcal{D}_f$.

\subsection{Hard Dispersion Loss}
Building upon the notation introduced in Sec.~\ref{sec:dispersion}, let $\hat{x}_i$ denote the $\ell_2$-normalized embedding of sample $i$ from an unlearning batch of size $B$, sampled from $\mathcal{D}_f$. We use the same definitions of $\mathcal{N}_{\text{pos}}$ (the set of anchors with at least one positive sample in the batch) and $P_i$ (the set of positive samples of anchor $i$). We minimize the following loss function:

\begin{equation}
\mathcal{L}_{\text{hard-disp}} =
\frac{1}{|\mathcal{N}_{\text{pos}}|}
\sum_{\substack{i \in \mathcal{N}_{\text{pos}}}}^{B}
\max\big(0,\, m + \max_{j \in P_i} (\hat{x}_i^\top \hat{x}_j)\big),
\label{eq:hard-dispersion}
\end{equation}
where $m$ is the margin. The operator $\max_{j \in P_i}$ selects the hardest positive for each anchor, i.e., the positive sample with the highest $\cos\theta$ to $\hat{x}_i$. This formulation differs from Eq.~\ref{eq:dispersion} in that it targets only the closest (i.e., hardest) positive pairs within each compact cluster, rather than uniformly pushing apart all positive representations. By penalizing only the most similar positives, the loss specifically aims at the strongest within-cluster ties, thereby more effectively disrupting cluster compactness and further degrading retrieval performance.

\subsection{Baseline Adaptations}
Since several baselines were originally designed with the Cross-Entropy loss, and are therefore not directly compatible with our encoder, we introduce minor adaptations of these methods to the CosFace~\cite{Wang2018} formulation.

For Random Labeling (RL)~\cite{golatkar2020eternal, hayase2020selective}, the adapted loss becomes:
\begin{equation}
\mathcal{L}_{\text{RL}} = 
\mathcal{L}_{\text{CosFace}}(f_{\omega}(I_i), y_{\text{rand}}) +
\lambda_{\text{retain}} \, \mathcal{L}_{\text{CosFace}}(f_{\omega}(I_j), y_j),
\label{eq:rl_adaptation}
\end{equation}
where $y_{\text{rand}} \sim \mathcal{Y}$ is drawn from a \textit{uniform} distribution over the label space, $I_i \sim \mathcal{D}_f$ is an image from the forget set, and $(I_j, y_j) \in \mathcal{D}_r$ is a retained sample. Since RL is known to be an aggressive forgetting strategy, the coefficient $\lambda_{\text{retain}}$ controls the amount of retained information required to prevent excessive forgetting.

For Gradient Ascent (GA)~\cite{golatkar2020eternal}, the adapted loss becomes:
\begin{equation}
\mathcal{L}_{\text{GA}} = -\,\mathcal{L}_{\text{CosFace}}(f_{\omega}(I_i), y_i),
\label{eq:ga_adaptation}
\end{equation}
where $(I_i, y_i) \in \mathcal{D}_f$ is a sample from the forget set.

For Boundary Shrink (BS)~\cite{chen2023boundary}, the adapted loss becomes:
\begin{equation}
\mathcal{L}_{\text{BS}} =
\mathcal{L}_{\text{CosFace}}(f_{\omega}(I_i), y_{\text{adv}}),
\label{eq:bs_adaptation}
\end{equation}
subject to
\[
\begin{aligned}
I_i' &= I_i + \epsilon \cdot \mathrm{sign}(\nabla_I \mathcal{L}_{\text{CosFace}}(f_{\omega}(I_i), y_i)),\\ 
y_{\text{adv}} &= 
\arg\max_{k \in \mathcal{Y}}
\hat{W}_k^\top \hat{x}_i',
\end{aligned}
\]
where $(I_i, y_i) \in \mathcal{D}_f$ is a sample from the forget set, $\hat{W}_k$ and $\hat{x}_i'$ follow Eq.~\ref{eq:cosine} and denote the $\ell_2$-normalized versions of the weight vector of the $k$-th class and the embedding of the adversarial image $I_i'$, respectively.

\section{Experiments}
\subsection{Datasets}
We conduct experiments using the following datasets: CelebA~\cite{liu2015faceattributes}, VGGFace2~\cite{cao2018vggface2}, CFP~\cite{cfp-paper}, and IJB-B~\cite{ijbb2017}. CelebA contains 202{,}599 images of 10{,}177 identities.  VGGFace2 contains 3.31 million images of 9{,}131 identities, with an average of 362.6 images per identity. CFP contains roughly 7{,}000 images of 500 identities; each identity has 10 frontal-view and 4 profile-view images, yielding 5{,}000 frontal and 2{,}000 profile images. IJB-B contains 21{,}798 still images and 55{,}026 frames extracted from 7{,}011 videos of 1{,}845 identities.

\subsection{Experimental Setup}
\label{subsec:exp_setup}
We structure our experimental setup into three groups of benchmarks corresponding to the forgetting scenarios and retention tests described below.

\textit{CelebA forgetting benchmarks.} For the CelebA forgetting scenario, the dataset is split identity-wise into a forget set $\mathcal{D}_f$ and a retain set $\mathcal{D}_r$. Unlearning is performed on $\mathcal{D}_f^{\text{train}}$, while the test split $\mathcal{D}_f^{\text{test}}$ provides query images for evaluating forgetting performance. We consider two versions of this benchmark: (i) \textit{base}, where retrieval is conducted with the gallery defined as $\text{CFP} \cup \mathcal{D}_f^{\text{test}}$ (with exact self-matches removed), and (ii) \textit{extended}, where the gallery is further augmented with 75{,}000 distractor images from IJB-B (still images and video frames).

\textit{CelebA retention benchmarks.} Retention performance is evaluated on two standard benchmarks: CFP-FP (frontal-to-profile search) and the VGGFace2 test split (500 identities, each divided into query and gallery subsets). Both benchmarks are reported in their \textit{base} form, i.e., with their native galleries. For the extended setting, we augment the CFP-FP gallery with the same 75{,}000 IJB-B distractors. We do not construct an extended version of the VGGFace2 test benchmark, as it already features a large gallery and query set, and the extended variant would be prohibitively expensive while offering limited additional insight.

\textit{VGGFace2 forgetting scenario (transferability test).} To assess transferability of our unlearning method, we additionally consider a forgetting scenario on VGGFace2. We sample an identity subset to form $\mathcal{D}_f$, split the images of each identity into train/test subsets, perform unlearning on $\mathcal{D}_f^{\text{train}}$, and evaluate forgetting using the \textit{extended} retrieval protocol (queries $= \mathcal{D}_f^{\text{test}}$, gallery $= \text{CFP} \cup \mathcal{D}_f^{\text{test}} \cup \text{IJB-B}^{\text{75k}}$). Retention in this transfer setting is assessed only via the extended CFP-FP benchmark.

To avoid any identity leakage from the forget identities into the evaluation galleries, we verify that the selected forget sets (for both CelebA and VGGFace2 scenarios) are identity-disjoint from all distractor datasets used in retrieval (CFP and IJB-B). We extract embeddings using a strong face recognition model (\textit{InsightFace}), compute class centroids for the forget identities and for CFP/IJB-B subjects, and evaluate cross-set cosine similarities. No centroid-level matches or high-similarity pairs were observed, confirming that the sets are fully disjoint. We also manually inspected the top-ranked cross-set matches and found no evidence of identity overlap.

\subsection{Evaluation Metrics}
For the retrieval benchmarks, we adopt the widely used Cumulative
Matching Characteristic (CMC) at Rank-1 (R@1) and mean Average Precision
(mAP) to evaluate both forgetting and retention performance.

For the standard class unlearning metrics, we report centroid-level accuracy on $\mathcal{D}_f^{\text{test}}$ and $\mathcal{D}_r^{\text{test}}$. To quantitatively assess the compactness of clusters corresponding to forgotten and retained identities, we compute the Cluster Compactness Score (CS), defined as the average pairwise cosine similarity within each identity cluster, calculated as follows:

\begin{equation}
CS = \frac{1}{|\mathcal{P}|} \sum_{p \in \mathcal{P}} \left( \frac{1}{n_p(n_p-1)} \sum_{i=1}^{n_p} \sum_{\substack{j=1 \\ j \neq i}}^{n_p} \hat{x}_i^\top \hat{x}_j \right),
\label{eq:cs_metric}
\end{equation}

where $\mathcal{P} \subset \mathcal{Y}$ is a subset of identities (either $\mathcal{P}_u$ or $\mathcal{P}_r$, where $\mathcal{P}_r$ denotes the retained identities),  $p \in \mathcal{P}$ is an identity label, and $\{\hat{x}_i\}_{i=1}^{n_p}$ denotes the set of $\ell_2$-normalized embeddings of all images belonging to identity $p$, as defined in Eq.~\ref{eq:cosine}.

All reported metrics are averaged over three independently sampled forget sets, and we report mean ± standard deviation across these runs.

\begin{table*}[t]
\centering
\small
\setlength{\tabcolsep}{8pt}
\renewcommand{\arraystretch}{1.15}
\begin{tabular}{l|cc|cc|cc}
\hline
Methods &
\multicolumn{2}{c|}{$\mathcal{D}_f^{\text{test}}$} &
\multicolumn{2}{c|}{CFP-FP} &
\multicolumn{2}{c}{VGGFace2} \\
\cline{2-7}
& mAP & R@1 & mAP & R@1 & mAP & R@1 \\
\hline

Original &
88.70 $\pm$ 1.00 & 98.17 $\pm$ 1.01 &
91.00 $\pm$ 0.00 & 97.50 $\pm$ 0.00 &
89.10 $\pm$ 0.00 & 98.80 $\pm$ 0.00 \\

\hline

Random Labeling &
87.73 $\pm$ 1.50 & 98.17 $\pm$ 1.01 &
90.00 $\pm$ 0.36 & 97.33 $\pm$ 0.06 &
88.40 $\pm$ 0.10 & 98.80 $\pm$ 0.00 \\

Gradient Ascent &
86.80 $\pm$ 0.87 & 98.17 $\pm$ 1.01 &
89.40 $\pm$ 0.66 & 97.20 $\pm$ 0.20 &
88.03 $\pm$ 0.45 & 98.80 $\pm$ 0.00 \\

Lipschitz Unlearning &
84.57 $\pm$ 2.69 & 97.93 $\pm$ 1.39 &
87.67 $\pm$ 1.59 & 97.07 $\pm$ 0.32 &
86.17 $\pm$ 1.23 & 98.73 $\pm$ 0.06 \\

Contrastive Unlearning &
88.57 $\pm$ 0.85 & 98.17 $\pm$ 1.01 &
91.00 $\pm$ 0.00 & 97.57 $\pm$ 0.06 &
89.00 $\pm$ 0.00 & 98.80 $\pm$ 0.00 \\

Boundary Shrink &
60.23 $\pm$ 5.06 & 96.53 $\pm$ 0.81 &
88.47 $\pm$ 0.25 & 97.13 $\pm$ 0.15 &
87.03 $\pm$ 0.31 & 98.73 $\pm$ 0.06 \\

\rowcolor{gray!12}
Dispersion Loss &
17.83 $\pm$ 1.51 & 67.37 $\pm$ 9.35 &
89.20 $\pm$ 0.44 & 97.30 $\pm$ 0.10 &
87.50 $\pm$ 0.26 & 98.80 $\pm$ 0.00 \\

\rowcolor{gray!12}
Hard Dispersion Loss &
16.00 $\pm$ 1.87 & 65.87 $\pm$ 7.57 &
89.20 $\pm$ 0.36 & 97.27 $\pm$ 0.12 &
87.47 $\pm$ 0.23 & 98.80 $\pm$ 0.00 \\

\hline
\end{tabular}

\caption{Evaluation of mAP and R@1 on the base CelebA forgetting and retention retrieval benchmarks.}
\label{tab:celeba_base}
\end{table*}

\begin{table}[t]
\centering
\small
\setlength{\tabcolsep}{4pt}
\renewcommand{\arraystretch}{1.15}
\resizebox{\columnwidth}{!}{
\begin{tabular}{l|cc|cc}
\hline
Methods &
\multicolumn{2}{c|}{$\mathcal{D}_f^{\text{test}}$} &
\multicolumn{2}{c}{$\mathcal{D}_r^{\text{test}}$} \\
\cline{2-5}
& Accuracy & CS & Accuracy & CS \\
\hline

Original & 98.20 $\pm$ 0.95 & 0.615 $\pm$ 0.025 & 96.60 $\pm$ 0.00 & 0.616 $\pm$ 0.000 \\
\hline
RL & 0.00 $\pm$ 0.00 & 0.615 $\pm$ 0.026 & 96.50 $\pm$ 0.00 & 0.626 $\pm$ 0.003 \\
GA & 0.57 $\pm$ 0.49 & 0.605 $\pm$ 0.067 & 96.33 $\pm$ 0.12 & 0.659 $\pm$ 0.005 \\
LU & 
56.63 $\pm$ 12.50 & 0.561 $\pm$ 0.040 & 96.33 $\pm$ 0.12 & 0.637 $\pm$ 0.016 \\
CU & 98.17 $\pm$ 1.01 & 0.581 $\pm$ 0.024 & 96.60 $\pm$ 0.00 & 0.620 $\pm$ 0.001 \\
BS & 5.33 $\pm$ 4.14 & 0.255 $\pm$ 0.013 & 96.37 $\pm$ 0.06 & 0.594 $\pm$ 0.003 \\
\rowcolor{gray!12}
Disp & 
0.80 $\pm$ 0.85 & 0.089 $\pm$ 0.012 & 96.43 $\pm$ 0.12 & 0.599 $\pm$ 0.004 \\
\rowcolor{gray!12}
HardDisp & 
1.33 $\pm$ 0.40 & 0.086 $\pm$ 0.010 & 96.47 $\pm$ 0.06 & 0.597 $\pm$ 0.003 \\
\hline

\end{tabular}
}
\caption{Evaluation of centroid-level accuracy and cluster compactness score (CS) for the CelebA forgetting scenario. Method names follow the abbreviated notation and ordering of Table~\ref{tab:celeba_base}.}
\label{tab:acc_cs_table}
\end{table}

\subsection{Baselines}
We compare our method against the following unlearning baselines:

\noindent \textit{Random Labeling (RL)}~\cite{golatkar2020eternal, hayase2020selective}: we use our adapted formulation in Eq.~\ref{eq:rl_adaptation}, finetuning the original model on the randomly relabeled forget set $\mathcal{D}_f^{\text{train}}$.

\noindent \textit{Gradient Ascent (GA)}~\cite{golatkar2020eternal}: we use the adaptation in Eq.~\ref{eq:ga_adaptation}, finetuning the original model on $\mathcal{D}_f^{\text{train}}$ in the direction of gradient ascent.

\noindent \textit{Contrastive Unlearning (CU)}~\cite{lee2025contrastive}: we finetune the original model by contrasting embeddings from the forget set $\mathcal{D}_f^{\text{train}}$ against those in the retain set $\mathcal{D}_r^{\text{train}}$.

\noindent \textit{Lipschitz Unlearning (LU)}~\cite{foster2024zeroshot, kravets2025zeroshot}: we finetune the original model on $\mathcal{D}_f^{\text{train}}$ with Lipschitz regularization, encouraging smoother embedding geometry for the forgotten identities, thus weakening identity-specific information.

\noindent \textit{Boundary Shrink (BS)}~\cite{chen2023boundary}: we use the adaptation in Eq.~\ref{eq:bs_adaptation}, finetuning the model on $\mathcal{D}_f^{\text{train}}$ with adversarial labels derived via an FGSM attack.

\subsection{Implementation Details}
\label{subsec:impl_details}
We use an IResNet-50~\cite{he2016resnet} encoder initialized with \textit{InsightFace}~\cite{insightface2019} pretrained weights (Glint360K~\cite{an2021partial}, CosFace~\cite{Wang2018}). We first fine-tune this model on the CelebA training split $(\mathcal{D}_f^{\text{train}} \cup \mathcal{D}_r^{\text{train}})$ using the CosFace loss (margin $m=0.4$, scale $s=64$), producing 512-dimensional $\ell_2$-normalized embeddings. Fine-tuning is performed with SGD (learning rate $1\times10^{-3}$, momentum 0.9, weight decay $5\times10^{-4}$), batch size 128, and a linear learning-rate decay over 10 epochs.

All unlearning methods start from this shared model. We train the full backbone with SGD and random horizontal flipping as the only augmentation. Our dispersion-based methods use a learning rate $1\times10^{-4}$, margin $m=0.2$, batch size 32 (160 for VGGFace2 experiments), and are trained for 1{,}000 iterations. No retention loss is used ($\lambda_{\text{retain}}=0$).

Baseline methods use the same optimizer and augmentation settings for fairness. The only differences are as follows: Boundary Shrink and Gradient Ascent use a lower learning rate $1\times10^{-5}$; Lipschitz Unlearning uses a retention weight $\lambda_{\text{retain}} = 0.05$ and the SalUn-0.5~\cite{fan2024salun} weight saliency estimation technique; and Random Labeling uses $\lambda_{\text{retain}} = 1.0$. All other baselines use $\lambda_{\text{retain}} = 0$.

Each baseline is reported with the configuration that achieved the best trade-off between forgetting and retention, including its training length. Additional experimental results for alternative settings, including failure cases, are provided in the Appendix~\ref{sec:baseline_add_info}. All experiments are conducted on a single NVIDIA A40 (48GB) using the PyTorch toolkit~\cite{paszke2019pytorch}.

\subsection{Results}
For all forgetting scenarios, we randomly sample 10 identities to form $\mathcal{P}_u$ and construct $\mathcal{D}_f$ / $\mathcal{D}_r$ and their splits as described in Sec.~\ref{subsec:exp_setup}. In CelebA, this yields approximately 15 images per forgotten identity in both $\mathcal{D}_f^{\text{train}}$ and $\mathcal{D}_f^{\text{test}}$, whereas in VGGFace2 the same procedure yields approximately 145 images per forgotten identity per split.

\begin{table*}[t]
\centering
\small
\setlength{\tabcolsep}{8pt}
\renewcommand{\arraystretch}{1.15}

\begin{tabular}{l|cc|cc}
\hline
Methods &
\multicolumn{2}{c|}{$\mathcal{D}_f^{\text{test}}$} &
\multicolumn{2}{c}{CFP-FP} \\
\cline{2-5}
& mAP & R@1 & mAP & R@1 \\
\hline

Original &
88.40 $\pm$ 0.80 & 98.17 $\pm$ 1.01 &
70.10 $\pm$ 0.00 & 74.10 $\pm$ 0.00 \\
\hline

Random Labeling &
85.87 $\pm$ 2.06 & 98.17 $\pm$ 1.01 &
68.27 $\pm$ 0.47 & 73.70 $\pm$ 0.20 \\

Gradient Ascent &
84.27 $\pm$ 0.91 & 97.90 $\pm$ 0.69 &
67.27 $\pm$ 1.23 & 73.53 $\pm$ 0.50 \\

Lipschitz Unlearning &
82.33 $\pm$ 2.49 & 97.67 $\pm$ 1.10 &
63.80 $\pm$ 2.54 & 72.80 $\pm$ 0.85 \\

Contrastive Unlearning &
88.23 $\pm$ 0.70 & 98.17 $\pm$ 1.01 &
70.03 $\pm$ 0.06 & 74.00 $\pm$ 0.10 \\

Boundary Shrink &
49.70 $\pm$ 3.99 & 92.50 $\pm$ 0.70 &
65.70 $\pm$ 0.52 & 73.17 $\pm$ 0.12 \\

\rowcolor{gray!12}
Dispersion Loss &
13.03 $\pm$ 1.19 & 57.37 $\pm$ 7.44 &
67.03 $\pm$ 0.83 & 73.47 $\pm$ 0.21 \\

\rowcolor{gray!12}
Hard Dispersion Loss &
11.53 $\pm$ 1.46 & 56.33 $\pm$ 5.84 &
67.03 $\pm$ 0.74 & 73.43 $\pm$ 0.23 \\
\hline

\end{tabular}

\caption{Evaluation of mAP and R@1 on the extended CelebA forgetting and retention retrieval benchmarks.}
\label{tab:celeba_ext}
\end{table*}

\begin{table}[t]
\centering
\small
\setlength{\tabcolsep}{4pt}
\renewcommand{\arraystretch}{1.15}

\resizebox{\columnwidth}{!}{
\begin{tabular}{l|cc|cc}
\hline
Methods &
\multicolumn{2}{c|}{$\mathcal{D}_f^{\text{test}}$} &
\multicolumn{2}{c}{CFP-FP} \\
\cline{2-5}
& mAP & R@1 & mAP & R@1 \\
\hline

Original &
89.40 $\pm$ 10.00 & 97.93 $\pm$ 1.85 &
70.10 $\pm$ 0.00 & 74.10 $\pm$ 0.00 \\
\hline

Disp &
4.70 $\pm$ 0.61 & 79.03 $\pm$ 6.22 &
63.23 $\pm$ 2.14 & 72.17 $\pm$ 0.74 \\

HardDisp &
1.53 $\pm$ 0.21 & 51.60 $\pm$ 4.29 &
57.30 $\pm$ 3.87 & 69.70 $\pm$ 1.97 \\
\hline

\end{tabular}
}
\caption{Transferability test: mAP and R@1 on the VGGFace2 forgetting and retention retrieval benchmarks.}
\label{tab:vggface2_ext}
\end{table}

\smallskip
\textbf{CelebA Forgetting Scenario.} Results for the standard class unlearning metrics are presented in Table~\ref{tab:acc_cs_table}. The analysis shows that several baselines (RL, GA, BS) obtain strong forgetting performance when evaluated by centroid-level accuracy. However, for RL and GA this effect does not transfer to the compactness score (CS), indicating that they do not substantially affect the internal structure of identity clusters. Boundary Shrink demonstrates the opposite behavior: although it forgets less according to centroid accuracy, it yields a lower CS, indicating a stronger disruption of cluster compactness. In contrast, our method matches the best baselines (RL and GA) in terms of forgetting and retention accuracy, while achieving the strongest effect on CS, indicating the largest disruption of cluster compactness among all compared approaches.

Results for the base CelebA forgetting and retention retrieval benchmarks are reported in Table~\ref{tab:celeba_base}. Forgetting performance closely follows the behavior of the cluster compactness score (CS): methods that fail to disrupt cluster structure also show poor forgetting under both mAP and R@1. Most baselines show only limited forgetting, with Boundary Shrink being the strongest among them. Our method exhibits a substantial improvement in forgetting, yielding $\Delta_\text{mAP} = 44.23$ and $\Delta_\text{R@1} = 30.66$ under the retrieval evaluation compared to Boundary Shrink. Meanwhile, it maintains stronger retention performance across both retention benchmarks.

We further evaluate forgetting and retention performance under the expanded gallery setup described in Sec.~\ref{subsec:exp_setup}. Results for the extended retrieval benchmarks are presented in Table~\ref{tab:celeba_ext}. Increasing the number of distractors creates a more challenging retrieval scenario, allowing for a stricter assessment of both forgetting and retention. The overall trends remain consistent with the base setting, with Boundary Shrink being the strongest baseline. Our method outperforms Boundary Shrink by $\Delta_\text{mAP} = 38.17$ and $\Delta_\text{R@1} = 36.17$, while still maintaining stronger retention performance.

\smallskip
\textbf{VGGFace2 Forgetting Scenario.} To evaluate the transferability of our approach, we additionally run unlearning experiments on VGGFace2. The method is applied without any additional tuning; the only adjustment is increasing the batch size to 160 as stated in Sec.~\ref{subsec:impl_details}. This scaling preserves the order of magnitude of the expected fraction of an identity’s images seen as positives per parameter update, which would otherwise drop significantly with the default sampler due to a much larger per-identity image count. Results are reported in Table~\ref{tab:vggface2_ext}. The overall unlearning behavior remains consistent with our previous analysis, yielding similar forgetting performance in terms of mAP and R@1. However, we observe a slightly larger drop in retention performance, which we attribute to the need for retention losses and hyperparameters tuning when operating on the substantially larger number of images per identity.

\begin{figure*}[t]
    \centering
    \begin{subfigure}{0.24\linewidth}
        \centering
        \includegraphics[width=\linewidth]{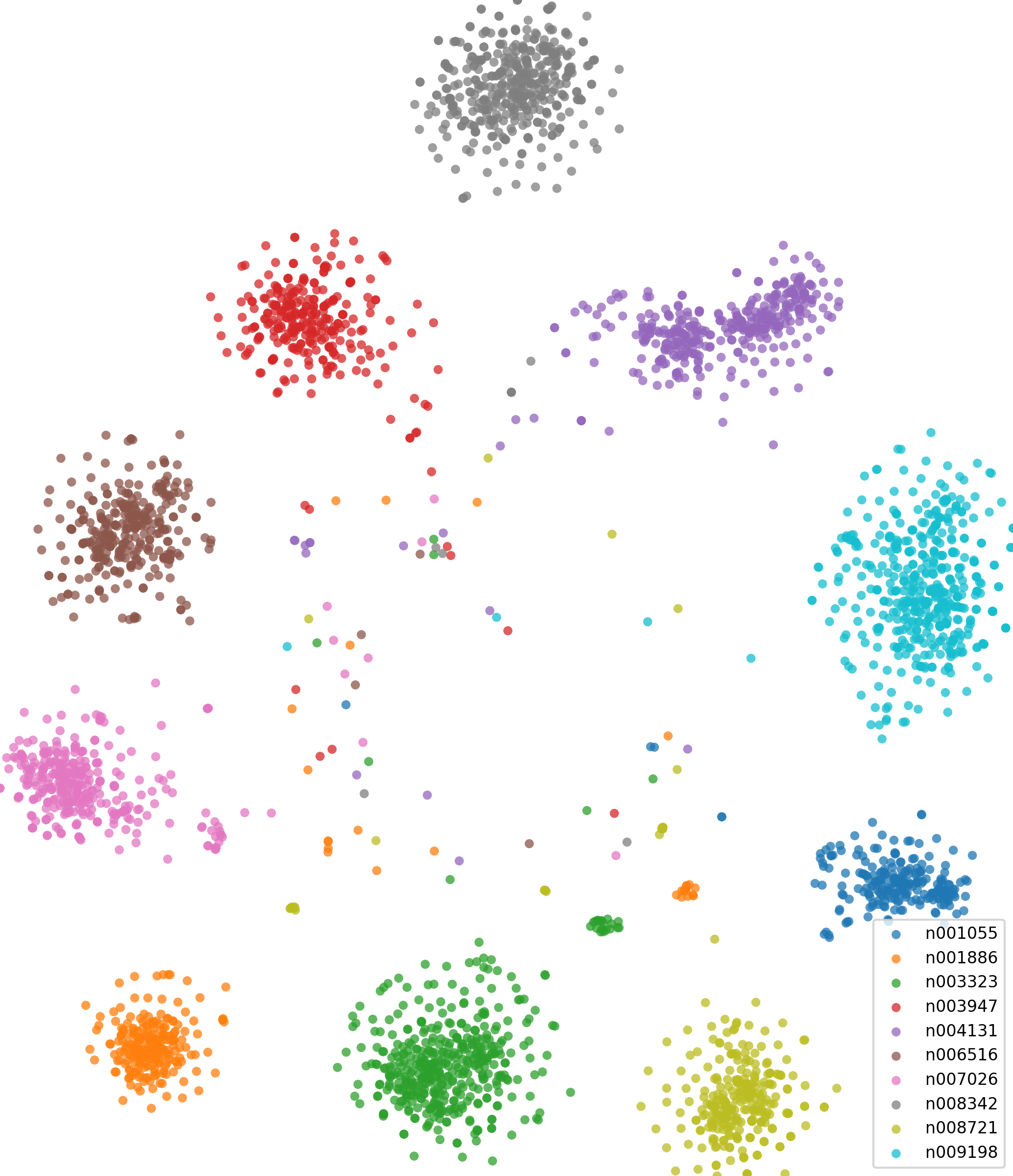}
        \caption{$\mathcal{D}_f$ Before}
        \label{fig:df_before}
    \end{subfigure}
    \begin{subfigure}{0.24\linewidth}
        \centering
        \includegraphics[width=\linewidth]{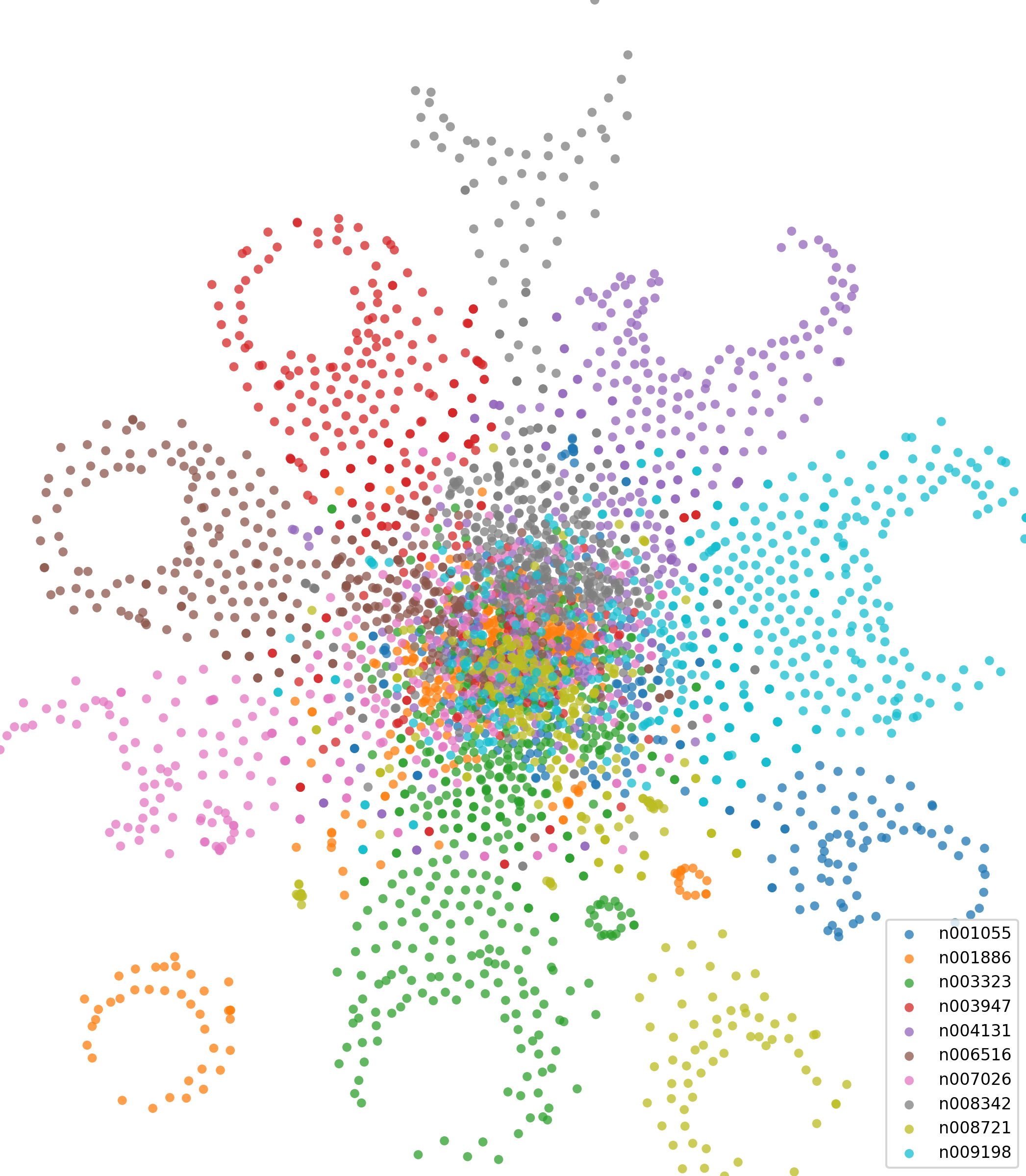}
        \caption{$\mathcal{D}_f$ After}
        \label{fig:df_after}
    \end{subfigure}
    \begin{subfigure}{0.24\linewidth}
        \centering
        \includegraphics[width=\linewidth]{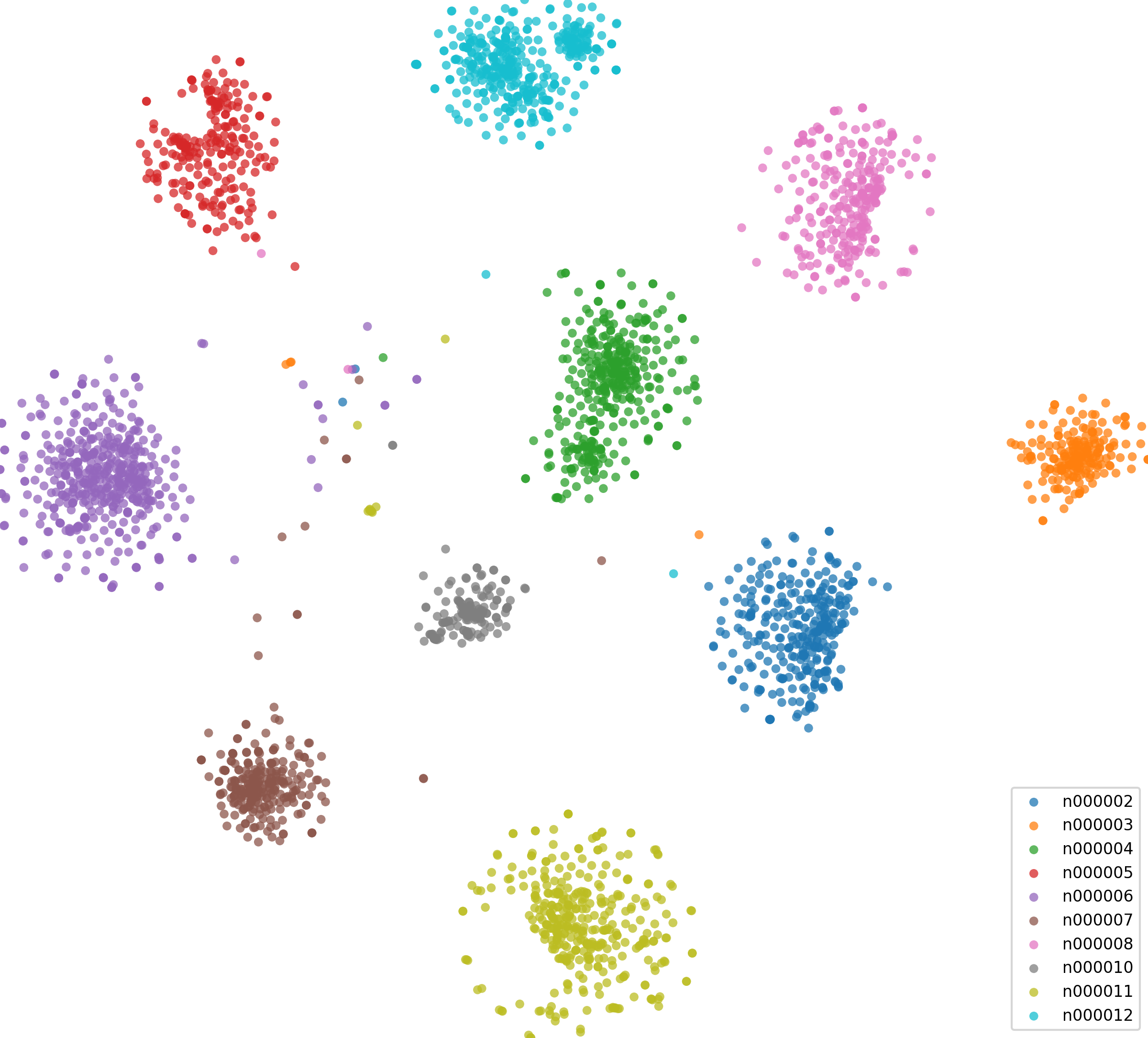}
        \caption{$\mathcal{D}_r$ Before}
        \label{fig:dr_before}
    \end{subfigure}
    \begin{subfigure}{0.24\linewidth}
        \centering
        \includegraphics[width=\linewidth]{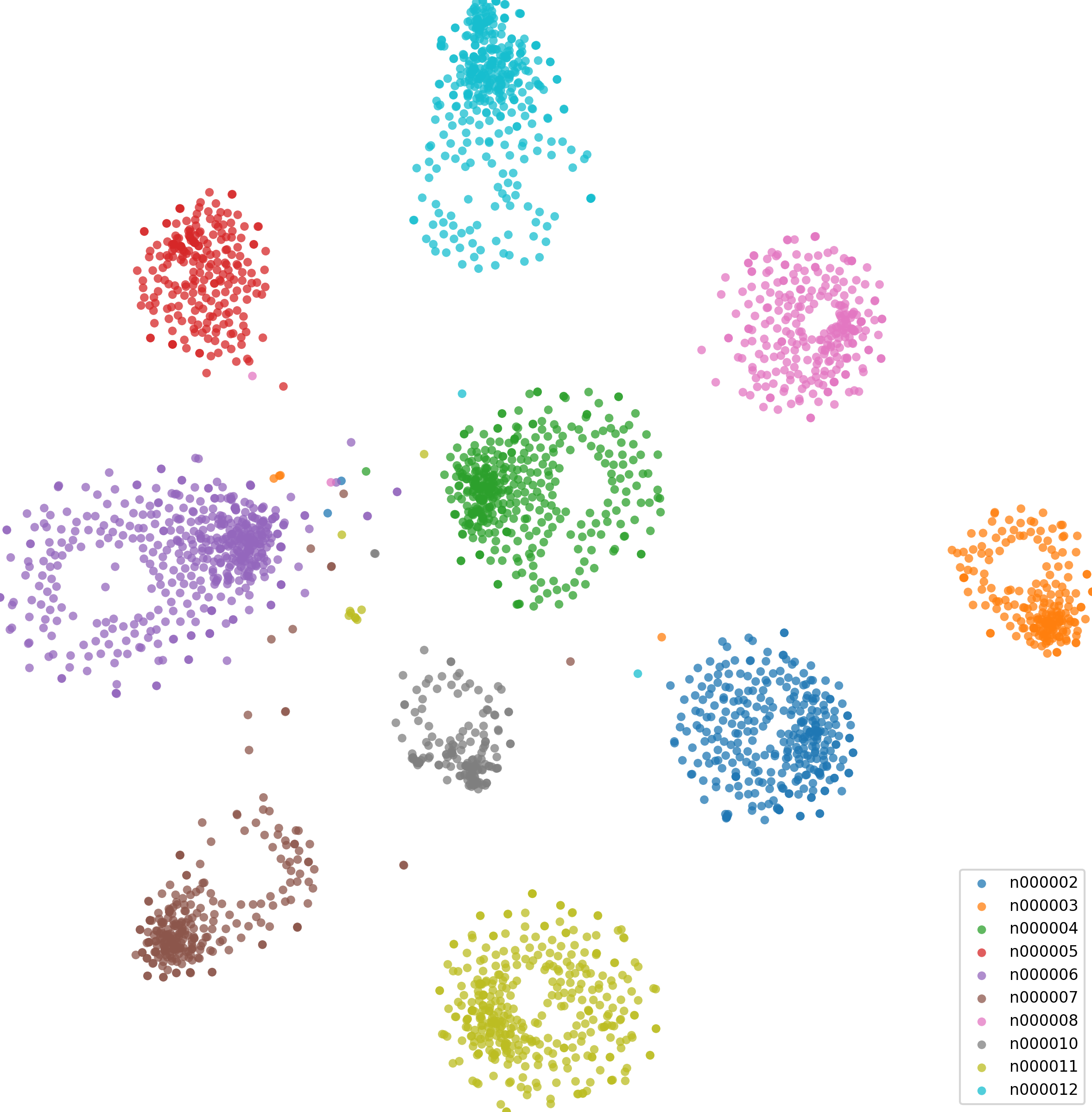}
        \caption{$\mathcal{D}_r$ After}
        \label{fig:dr_after}
    \end{subfigure}
    \caption{t-SNE visualizations of the raw embeddings before and after unlearning for the forget set and a random subset of 10 identities from the retain set on VGGFace2.}
    \label{fig:vis_embeddings}
\end{figure*}

We visualize the raw embeddings using t-SNE~\cite{tsne} for $\mathcal{D}_f$ and for a randomly selected subset of $\mathcal{D}_r$ before and after unlearning (Figure~\ref{fig:vis_embeddings}). Before unlearning, the forget-set identities form compact and well-separated clusters (Figure~\ref{fig:df_before}). After unlearning, this structure largely disappears: most embeddings collapse into overlapping, unstructured regions, while a small fraction still forms weakly separable groups, consistent with the remaining nonzero R@1 (Figure~\ref{fig:df_after}). The retain identities demonstrate clear, compact clusters prior to unlearning (Figure~\ref{fig:dr_before}), and although some clusters become slightly less dense afterward, their overall separability and structure remain largely preserved (Figure~\ref{fig:dr_after}).

\section{Limitations and Discussion}
\label{sec:limitations}

We observe a notable discrepancy between mAP and R@1 in the forgetting setting: while mAP collapses sharply after unlearning ($\Delta_\text{mAP} \approx 73\text{--}88$ relative to the original model), R@1 decreases much more moderately ($\Delta_\text{R@1} \approx 32\text{--}46$), which suggests that the cluster compactness score correlates strongly with mAP but less strongly with R@1. Another consistent observation is that different identities are forgotten to different extents, and this variation does not correlate with dataset statistics such as the number of images per identity. Instead, the effect might be tied to how the base model encodes particular subjects, suggesting that “forgettability” might be an identity-dependent property to some extent. Finally, despite substantial improvements over the baselines and a drastic drop in mAP, the absolute R@1 values for forgotten identities remain around $51$--$57\%$. This implies that the nearest neighbor of a query often retains the same identity label. This is partly explained by the presence of near-duplicate images in the dataset and partly by the inherently low intra-class variability of aligned facial images. Potential extensions may include SimCLR~\cite{chen2020simclr}-style dispersion, where augmented views of the same image are explicitly pushed apart, possibly combined with retain losses to achieve a more controlled variant of aggressive unlearning.

\section{Conclusion}
In this paper, we studied the problem of face identity unlearning in deep embedding models, highlighting its unique challenges compared to classification-based unlearning. Our evaluation of existing approximate unlearning methods revealed that they struggle to disrupt the embedding-level identity structure, showing poor forgetting under a retrieval scenario. To address this, we proposed a dispersion-based approach for unlearning that explicitly disperses compact identity clusters. Extensive experiments across multiple datasets and benchmarks demonstrate that this simple formulation consistently outperforms baseline unlearning algorithms. 

{\small
\bibliographystyle{ieee_fullname}
\bibliography{egbib}

@String(AAAI = {AAAI Conference on Artificial Intelligence})

@String(CVPR = {IEEE Conference on Computer Vision and Pattern Recognition})

@String(ICCV = {International Conference on Computer Vision})

@String(ICML = {International Conference on Machine Learning})

@String(IJCAI = {International Joint Conference on Artificial Intelligence})

@String(ECCV = {European Conference on Computer Vision})

@String(FG = {International Conference on Automatic Face and Gesture Recognition})

@String(ICLR = {International Conference on Learning Representations})

@String(CVPRW= {IEEE Conference on Computer Vision and Pattern Recognition Workshop})

@inproceedings{taigman2014deepface,
  title={{DeepFace}: Closing the gap to human-level performance in face verification},
  author={Taigman, Yaniv and Yang, Ming and Ranzato, Marc'Aurelio and Wolf, Lior},
  booktitle=CVPR,
  pages={1701--1708},
  year={2014}
}

@article{sun2015deepid3,
  title={{DeepID3}: Face recognition with very deep neural networks},
  author={Sun, Yi and Liang, Ding and Wang, Xiaogang and Tang, Xiaoou},
  journal={arXiv preprint arXiv:1502.00873},
  year={2015}
}

@inproceedings{schroff2015facenet,
  title={{FaceNet}: A unified embedding for face recognition and clustering},
  author={Schroff, Florian and Kalenichenko, Dmitry and Philbin, James},
  booktitle=CVPR,
  pages={815--823},
  year={2015}
}

@inproceedings{chopra2005learning,
  title={Learning a similarity metric discriminatively, with application to face verification},
  author={Chopra, Sumit and Hadsell, Raia and LeCun, Yann},
  booktitle=CVPR,
  volume={1},
  pages={539--546},
  year={2005},
  organization={IEEE}
}

@inproceedings{wang2017deep,
  title={Deep metric learning with angular loss},
  author={Wang, Jian and Zhou, Feng and Wen, Shilei and Liu, Xiao and Lin, Yuanqing},
  booktitle=ICCV,
  pages={2593--2601},
  year={2017}
}

@inproceedings{wen2016discriminative,
  title={A discriminative feature learning approach for deep face recognition},
  author={Wen, Yandong and Zhang, Kaipeng and Li, Zhifeng and Qiao, Yu},
  booktitle=ECCV,
  pages={499--515},
  year={2016},
  organization={Springer}
}

@article{Ranjan2017,
  title={L2-constrained softmax loss for discriminative face verification},
  author={Ranjan, Rajeev and Castillo, Carlos D and Chellappa, Rama},
  journal={arXiv preprint arXiv:1703.09507},
  year={2017}
}

@inproceedings{Liu2017,
  title={Sphereface: Deep hypersphere embedding for face recognition},
  author={Liu, Weiyang and Wen, Yandong and Yu, Zhiding and Li, Ming and Raj, Bhiksha and Song, Le},
  booktitle=CVPR,
  pages={212--220},
  year={2017}
}

@article{wang2018additive,
  title={Additive margin softmax for face verification},
  author={Wang, Feng and Cheng, Jian and Liu, Weiyang and Liu, Haijun},
  journal={IEEE Signal Processing Letters},
  volume={25},
  number={7},
  pages={926--930},
  year={2018},
  publisher={IEEE}
}

@inproceedings{Wang2018,
  title={{CosFace}: Large margin cosine loss for deep face recognition},
  author={Wang, Hao and Wang, Yitong and Zhou, Zheng and Ji, Xing and Gong, Dihong and Zhou, Jingchao and Li, Zhifeng and Liu, Wei},
  booktitle=CVPR,
  pages={5265--5274},
  year={2018}
}

@inproceedings{Deng2018,
  title={{ArcFace}: Additive angular margin loss for deep face recognition},
  author={Deng, Jiankang and Guo, Jia and Xue, Niannan and Zafeiriou, Stefanos},
  booktitle=CVPR,
  pages={4690--4699},
  year={2019}
}

@inproceedings{meng2021magface,
  title={{MagFace}: A universal representation for face recognition and quality assessment},
  author={Meng, Qiang and Zhao, Shichao and Huang, Zhida and Zhou, Feng},
  booktitle=CVPR,
  pages={14225--14234},
  year={2021}
}

@inproceedings{kim2022adaface,
  title={{AdaFace}: Quality adaptive margin for face recognition},
  author={Kim, Minchul and Choi, Anil and Kim, Donghyeon and Cho, Sunghyun and Kim, In So Kweon},
  booktitle=CVPR,
  pages={18750--18759},
  year={2022}
}

@inproceedings{huang2020curricularface,
  title={{CurricularFace}: Adaptive curriculum learning loss for deep face recognition},
  author={Huang, Yuge and Wang, Yang and Tai, Yu and Liu, Xiaoming and Shen, Chunhua and Li, Jie and Huang, Xiaogang},
  booktitle=CVPR,
  pages={5901--5910},
  year={2020}
}

@article{zhao2021facetransformer,
  title={{Face Transformer for Recognition}},
  author={Zhao, Shichao and Meng, Qiang and Chen, Yibo and Zhou, Feng},
  journal={arXiv preprint arXiv:2103.14803},
  year={2021}
}

@inproceedings{liu2015faceattributes,
  title = {Deep Learning Face Attributes in the Wild},
  author = {Liu, Ziwei and Luo, Ping and Wang, Xiaogang and Tang, Xiaoou},
  booktitle = {Proceedings of International Conference on Computer Vision (ICCV)},
  month = {December},
  year = {2015} 
}

@inproceedings{cao2018vggface2,
  title={{VGGFace2}: A dataset for recognising faces across pose and age},
  author={Cao, Qiong and Shen, Li and Xie, Weidi and Parkhi, Omkar M and Zisserman, Andrew},
  booktitle=FG,
  pages={67--74},
  year={2018},
  organization={IEEE}
}

@inproceedings{guo2016msceleb,
  title={{MS-Celeb-1M}: A dataset and benchmark for large-scale face recognition},
  author={Guo, Yandong and Zhang, Lei and Hu, Yuxiao and He, Xiaodong and Gao, Jianfeng},
  booktitle=ECCV,
  pages={87--102},
  year={2016},
  organization={Springer}
}

@inproceedings{he2016resnet,
  title={{Deep Residual Learning for Image Recognition}},
  author={He, Kaiming and Zhang, Xiangyu and Ren, Shaoqing and Sun, Jian},
  booktitle=CVPR,
  pages={770--778},
  year={2016}
}

@inproceedings{szegedy2015inception,
  title={{Going deeper with convolutions}},
  author={Szegedy, Christian and Liu, Wei and Jia, Yangqing and Sermanet, Pierre and Reed, Scott and Anguelov, Dragomir and Erhan, Dumitru and Vanhoucke, Vincent and Rabinovich, Andrew},
  booktitle=CVPR,
  pages={1--9},
  year={2015}
}

@inproceedings{simonyan2015vgg,
  title={{Very Deep Convolutional Networks for Large-Scale Image Recognition}},
  author={Simonyan, Karen and Zisserman, Andrew},
  booktitle=ICLR,
  year={2015}
}

@inproceedings{dosovitskiy2021vit,
  title={{An Image is Worth 16x16 Words: Transformers for Image Recognition at Scale}},
  author={Dosovitskiy, Alexey and Beyer, Lucas and Kolesnikov, Alexander and Weissenborn, Dirk and Zhai, Xiaohua and Unterthiner, Thomas and Dehghani, Mostafa and Minderer, Matthias and Heigold, Georg and Gelly, Sylvain and Uszkoreit, Jakob and Houlsby, Neil},
  booktitle=ICLR,
  year={2021}
}

@article{Hoofnagle02012019,
author = {Chris Jay Hoofnagle and Bart van der Sloot and Frederik Zuiderveen Borgesius},
title = {The European Union general data protection regulation: what it is and what it means*},
journal = {Information \& Communications Technology Law},
volume = {28},
number = {1},
pages = {65--98},
year = {2019},
publisher = {Routledge},
doi = {10.1080/13600834.2019.1573501}
}

@inproceedings{cao2015unlearning,
  title={{Towards Making Systems Forget with Machine Unlearning}},
  author={Yinzhi Cao and Junfeng Yang},
  booktitle={IEEE Symposium on Security and Privacy (S\&P)},
  pages={463--480},
  year={2015},
  publisher={IEEE}
}

@inproceedings{ginart2019making,
  title={{Making AI Forget You: Data Deletion in Machine Learning}},
  author={Antonio A. Ginart and Melody Y. Guan and Gregory Valiant and James Zou},
  booktitle={Advances in Neural Information Processing Systems (NeurIPS)},
  pages={3513--3526},
  year={2019}
}

@inproceedings{sekhari2021remember,
  title={{Remember What You Want to Forget: Algorithms for Machine Unlearning}},
  author={Ayush Sekhari and Jayadev Acharya and Gautam Kamath and Ananda Theertha Suresh},
  booktitle={Advances in Neural Information Processing Systems (NeurIPS)},
  year={2021}
}

@inproceedings{gupta2021adaptive,
  title={{Adaptive Machine Unlearning}},
  author={Varun Gupta and Christopher Jung and Seth Neel and Aaron Roth and Saeed Sharifi-Malvajerdi and Chris Waites},
  booktitle={Advances in Neural Information Processing Systems (NeurIPS)},
  year={2021}
}

@inproceedings{neel2021descent,
  title={{Descent-to-Delete: Gradient-Based Methods for Machine Unlearning}},
  author={Seth Neel and Aaron Roth and Saeed Sharifi-Malvajerdi},
  booktitle={Proceedings of the 32nd International Conference on Algorithmic Learning Theory (ALT)},
  year={2021}
}

@inproceedings{chien2022certified,
  title={{Certified Graph Unlearning}},
  author={Eli Chien and Chao Pan and Olgica Milenkovic},
  booktitle={NeurIPS 2022 New Frontiers in Graph Learning Workshop},
  year={2022}
}

@inproceedings{guo2020certified,
  title={{Certified Data Removal from Machine Learning Models}},
  author={Chuan Guo and Tom Goldstein and Awni Hannun and Laurens van der Maaten},
  booktitle={Proceedings of the 37th International Conference on Machine Learning (ICML)},
  year={2020}
}

@inproceedings{bourtoule2021machine,
  title={{Machine Unlearning}},
  author={Lukas Bourtoule and Varun Chandrasekaran and Christopher A. Choquette-Choo and Hengrui Jia and Adelin Travers and Baiwu Zhang and David Lie and Nicolas Papernot},
  booktitle={Proceedings of the 42nd IEEE Symposium on Security and Privacy (S\&P)},
  year={2021}
}

@inproceedings{thudi2022necessity,
  title={{On the Necessity of Auditable Algorithmic Definitions for Machine Unlearning}},
  author={Anvith Thudi and Hengrui Jia and Ilia Shumailov and Nicolas Papernot},
  booktitle={Proceedings of the 31st USENIX Security Symposium (USENIX Security 22)},
  year={2022}
}

@inproceedings{carlini2021membership,
  title={{Membership Inference Attacks From First Principles}},
  author={Nicholas Carlini and Steve Chien and Milad Nasr and Shuang Song and Andreas Terzis and Florian Tramèr},
  booktitle={IEEE Symposium on Security and Privacy (S\&P) 2022},
  year={2022}
}

@inproceedings{golatkar2020eternal,
  title={{Eternal Sunshine of the Spotless Net: Selective Forgetting in Deep Networks}},
  author={Aditya Golatkar and Alessandro Achille and Stefano Soatto},
  booktitle={Proceedings of the IEEE/CVF Conference on Computer Vision and Pattern Recognition (CVPR)},
  year={2020}
}

@inproceedings{kurmanji2023towards,
  title={{Towards Unbounded Machine Unlearning}},
  author={Meghdad Kurmanji and Peter Triantafillou and Jamie Hayes and Eleni Triantafillou},
  booktitle={Advances in Neural Information Processing Systems (NeurIPS)},
  year={2023}
}

@inproceedings{graves2021amnesiac,
  title={{Amnesiac Machine Learning}},
  author={Laura Graves and Vineel Nagisetty and Vijay Ganesh},
  booktitle={Proceedings of the AAAI Conference on Artificial Intelligence (AAAI)},
  year={2021}
}

@article{hayase2020selective,
  title={{Selective Forgetting of Deep Networks at a Finer Level than Samples}},
  author={Tomohiro Hayase and Suguru Yasutomi and Takashi Katoh},
  journal={arXiv preprint arXiv:2012.11849},
  year={2020}
}

@article{foster2024zeroshot,
  title={{Zero-Shot Machine Unlearning at Scale via Lipschitz Regularization}},
  author={Jack Foster and Kyle Fogarty and Stefan Schoepf and Cengiz Öztireli and Alexandra Brintrup},
  journal={arXiv preprint arXiv:2402.01401},
  year={2024}
}

@inproceedings{kravets2025zeroshot,
  title={{Zero-Shot Class Unlearning in CLIP with Synthetic Samples}},
  author={Alexey Kravets and Vinay P. Namboodiri},
  booktitle={Proceedings of the IEEE/CVF Winter Conference on Applications of Computer Vision (WACV)},
  year={2025}
}

@inproceedings{lee2025contrastive,
  title={{Contrastive Unlearning: A Contrastive Approach to Machine Unlearning}},
  author={Hong kyu Lee and Qiuchen Zhang and Carl Yang and Jian Lou and Li Xiong},
  booktitle={Proceedings of the Thirty-Fourth International Joint Conference on Artificial Intelligence (IJCAI)},
  year={2025}
}

@inproceedings{fan2024salun,
  title={{SalUn: Empowering Machine Unlearning via Gradient-based Weight Saliency in Both Image Classification and Generation}},
  author={Chongyu Fan and Jiancheng Liu and Yihua Zhang and Eric Wong and Dennis Wei and Sijia Liu},
  booktitle={International Conference on Learning Representations (ICLR)},
  year={2024}
}

@inproceedings{chen2023boundary,
  title={{Boundary Unlearning: Rapid Forgetting of Deep Networks via Shifting the Decision Boundary}},
  author={Min Chen and Weizhuo Gao and Gaoyang Liu and Kai Peng and Chen Wang},
  booktitle={Proceedings of the IEEE/CVF Conference on Computer Vision and Pattern Recognition (CVPR)},
  year={2023}
}

@inproceedings{an2021partial,
  title={{Partial FC: Training 10 Million Identities on a Single Machine}},
  author={Xiang An and Xuhan Zhu and Yang Xiao and Lan Wu and Ming Zhang and Yuan Gao and Bin Qin and Debing Zhang and Ying Fu},
  booktitle={2021 IEEE/CVF International Conference on Computer Vision Workshops (ICCVW)},
  year={2021}
}

@inproceedings{cfp-paper,
  author = {S. Sengupta and J.C. Cheng and C.D. Castillo and V.M. Patel and R.  Chellappa and D.W. Jacobs},
  booktitle = {IEEE Conference on Applications of Computer Vision},
  title = {Frontal to Profile Face Verification in the Wild},
  month = {February},
  year = {2016}
}

@inproceedings{ijbb2017,
  author={Whitelam, Cameron and Taborsky, Emma and Blanton, Austin and Maze, Brianna and Adams, Jocelyn and Miller, Tim and Kalka, Nathan and Jain, Anil K. and Duncan, James A. and Allen, Kristen and Cheney, Jordan and Grother, Patrick},
  booktitle={2017 IEEE Conference on Computer Vision and Pattern Recognition Workshops (CVPRW)}, 
  title={IARPA Janus Benchmark-B Face Dataset}, 
  year={2017}
}

@misc{insightface2019,
  title={InsightFace: 2D and 3D Face Analysis Library},
  author={Deng, Jiankang and Guo, Jia and Liu, Jing and Zhao, Hongwei and Zafeiriou, Stefanos},
  howpublished={\url{https://github.com/deepinsight/insightface}},
  note={Accessed: 2025-01},
  year={2019}
}

@inproceedings{paszke2019pytorch,
  title={PyTorch: An Imperative Style, High-Performance Deep Learning Library},
  author={Paszke, Adam and Gross, Sam and Massa, Francisco and Lerer, Adam and Bradbury, James and Chanan, Gregory and Killeen, Trevor and Lin, Zeming and Gimelshein, Natalia and Antiga, Luca and Desmaison, Alban and Kopf, Andreas and Yang, Edward and DeVito, Zachary and Raison, Martin and Tejani, Alykhan and Chilamkurthy, Sasank and Steiner, Benoit and Fang, Lu and Bai, Junjie and Chintala, Soumith},
  booktitle={Advances in Neural Information Processing Systems (NeurIPS)},
  year={2019}
}

@article{tsne,
  author={Laurens van der Maaten and Geoffrey Hinton},
  title={Visualizing Data using t-SNE},
  journal={Journal of Machine Learning Research},
  year={2008}
}

@inproceedings{chen2020simclr,
  title={A Simple Framework for Contrastive Learning of Visual Representations},
  author={Chen, Ting and Kornblith, Simon and Norouzi, Mohammad and Hinton, Geoffrey},
  booktitle={Proceedings of the 37th International Conference on Machine Learning (ICML)},
  year={2020}
}

@inproceedings{Peng_2025_CVPR,
    author={Peng, Yi-Xing and Tang, Yu-Ming and Lin, Kun-Yu and Yang, Qize and Meng, Jingke and Wei, Xihan and Zheng, Wei-Shi},
    title={Person De-reidentification: A Variation-guided Identity Shift Modeling},
    booktitle={Proceedings of the IEEE/CVF Conference on Computer Vision and Pattern Recognition (CVPR)},
    year={2025}
}
}

\clearpage
\appendix

\section{Additional Experimental Analysis}
\label{sec:baseline_add_info}

We conducted additional experiments for each baseline, performing a lightweight hyperparameter search aimed at achieving the best possible behavior. These experiments were carried out on a single CelebA~\cite{liu2015faceattributes} forget set, and the corresponding results are presented in Table~\ref{tab:celeba_hyp_search}. The configurations that demonstrated the best forgetting–retention trade-off were reported in the main paper. Note that the absolute metric values differ from those in the main paper, as the latter report mean ± std over three distinct forget sets, while the lightweight hyperparameter search is run on one split only. Nevertheless, the selected configurations fall within the expected range.

For Boundary Shrink (BS)~\cite{chen2023boundary}, the original paper uses a learning rate of $1\times10^{-5}$ and 10 unlearning epochs. In their face recognition setup, the authors unlearn a single class from VGGFace2~\cite{cao2018vggface2}, which has approximately 362.6 images per identity. While the batch size is not stated in the paper, the publicly released code uses a batch size of 64. Under this setting, the number of iterations per epoch is approximately $\lceil 362 / 64 \rceil = 6$, yielding an estimated upper bound of 60 unlearning iterations in total. Furthermore, a later study on Contrastive Unlearning~\cite{lee2025contrastive} reports that Boundary Shrink typically converges within 40--60 iterations when applied to a ResNet~\cite{he2016resnet}-based model. Motivated by these observations, we evaluate Boundary Shrink under 100 and 500 unlearning iterations, using learning rates of $1\times10^{-5}$ and $1\times10^{-4}$. In practice, this method demonstrates the strongest transferability to the retrieval scenario among all baselines, without breaking the model.

For Lipschitz Unlearning (LipLoss)~\cite{foster2024zeroshot}, the original paper uses a learning rate of $3\times10^{-4}$, a noise standard deviation of $0.5$, and a single unlearning epoch for full-class removal on a CNN-based model. The publicly available code sets the number of noise samples to $n = 25$. Motivated by these settings, we evaluate Lipschitz Unlearning under the following hyperparameter ranges: SalUn~\cite{fan2024salun} $\in \{0.1, 0.5\}$, $\text{std} \in \{0.1, 0.5\}$, $n = 25$, $\lambda_{\text{retain}} \in \{0.0, 0.05, 0.1, 0.5, 1.0\}$, $\text{iterations} \in \{5, 50, 500, 1000\}$, and a learning rate of $1\times10^{-4}$. For weight-saliency estimation, we experiment with the CosFace~\cite{Wang2018} Gradient Ascent~\cite{golatkar2020eternal} (CFGA), LipLoss, and EmbNorm (EN) (as used in the original implementation) loss functions. In practice, we observe that the method either breaks the model or yields only weak forgetting.

For Random Labeling (RL)~\cite{golatkar2020eternal, hayase2020selective}, we experiment with $\lambda_{\text{retain}} \in \{0.0, 0.1, 0.5, 1.0\}$, $\text{iterations} \in \{25, 250, 500\}$, and a learning rate of $1\times10^{-4}$. We observe that the method either does not translate to the retrieval scenario, yielding low $\Delta_\text{mAP}$ and $\Delta_\text{R@1}$ on $\mathcal{D}_f^{\text{test}}$, or breaks the model.

For Gradient Ascent (GA)~\cite{golatkar2020eternal}, we experiment with $\text{iterations} \in \{10, 25, 50\}$, and learning rates of $1\times10^{-4}$ and $1\times10^{-5}$. We observe that the method either does not transfer to the retrieval scenario, similar to RL, or breaks the model.

For Contrastive Unlearning, the original paper uses a learning rate of $1\times10^{-3}$ and $1\times10^{-4}$ for a ResNet-based model, and according to their graphs, it converges around 50--70 iterations. Motivated by these observations, we evaluate Contrastive Unlearning under the following hyperparameter ranges: $\tau \in \{0.1, 0.2, 0.5\}$, $\text{iterations} \in \{50, 100, 250\}$, $\lambda_{\text{retain}} \in \{0.0, 0.001, 0.05, 0.1\}$, and learning rates of $1\times10^{-3}$ and $1\times10^{-4}$. In practice, we observe that the method either breaks the model or yields only weak forgetting.

\begin{table*}[t]
\centering
\small
\setlength{\tabcolsep}{4pt}
\renewcommand{\arraystretch}{1.15}

\resizebox{\textwidth}{!}{
\begin{tabular}{l|ccccccc}
\hline
Methods & $\mathcal{D}_f^{\text{test}}$ mAP & $\mathcal{D}_f^{\text{test}}$ R@1 & CFP-FP R@1 & $\mathcal{D}_f^{\text{test}}$ CS & $\mathcal{D}_r^{\text{test}}$ CS & $\mathcal{D}_f^{\text{test}}$ Acc & $\mathcal{D}_r^{\text{test}}$ Acc \\
\hline
\textbf{Original} & \textbf{87.7} & \textbf{97.1} & \textbf{97.5} & \textbf{0.634} & \textbf{0.616} & \textbf{97.2} & \textbf{96.6} \\
Disp($m=0.2$) $lr=1\times10^{-4}$ $\lambda_{\text{retain}}=0.0$ 500 Iters & 20.7 & 75.4 & 97.3 & 0.086 & 0.596 & 3.5 & 96.5 \\
\hline
BS $lr=1\times10^{-5}$ $\lambda_{\text{retain}}=0.0$ HeadFreeze 100 Iters & 84.4 & 97.1 & 97.4 & 0.401 & 0.619 & 13.5 & 96.5 \\
BS $lr=1\times10^{-5}$ $\lambda_{\text{retain}}=0.0$ 100 Iters & 84.6 & 97.1 & 97.5 & 0.400 & 0.619 & 10.6 & 96.5 \\
\rowcolor{gray!12}
BS $lr=1\times10^{-5}$ $\lambda_{\text{retain}}=0.0$ 500 Iters & 63.5 & 96.4 & 97.0 & 0.265 & 0.592 & 10.1 & 96.4 \\
BS $lr=1\times10^{-4}$ $\lambda_{\text{retain}}=0.0$ 500 Iters & 72.3 & 95.7 & 96.9 & 0.345 & 0.567 & 10.6 & 96.3 \\
\hline
CFGA-SALUN0.5 + LipLoss($n=25$, $std=0.1$) $lr=1\times10^{-4}$ $\lambda_{\text{retain}}=0.0$ 500 Iters & 81.3 & 97.1 & 87.1 & 0.615 & 0.590 & 24.6 & 88.5 \\
CFGA-SALUN0.5 + LipLoss($n=25$, $std=0.1$) $lr=1\times10^{-4}$ $\lambda_{\text{retain}}=0.0$ 1000 Iters & 10.0 & 43.5 & 1.9 & 0.344 & 0.331 & 0.0 & 0.6 \\
CFGA-SALUN0.5 + LipLoss($n=25$, $std=0.5$) $lr=1\times10^{-4}$ $\lambda_{\text{retain}}=0.0$ 50 Iters & 38.3 & 84.1 & 7.7 & 0.786 & 0.812 & 2.2 & 3.7 \\
CFGA-SALUN0.1 + LipLoss($n=25$, $std=0.5$) $lr=1\times10^{-4}$ $\lambda_{\text{retain}}=0.0$ 50 Iters & 38.1 & 82.6 & 7.8 & 0.788 & 0.815 & 0.7 & 3.7 \\
LL-SALUN0.5 + LipLoss($n=25$, $std=0.1$) $lr=1\times10^{-4}$ $\lambda_{\text{retain}}=0.0$ 1000 Iters & 6.2 & 31.9 & 1.6 & 0.346 & 0.282 & 0.0 & 0.2 \\
LL-SALUN0.5 + LipLoss($n=25$, $std=0.5$) $lr=1\times10^{-4}$ $\lambda_{\text{retain}}=0.0$ 50 Iters & 37.6 & 81.9 & 7.6 & 0.791 & 0.818 & 1.4 & 3.4 \\
LL-SALUN0.1 + LipLoss($n=25$, $std=0.5$) $lr=1\times10^{-4}$ $\lambda_{\text{retain}}=0.0$ 50 Iters & 37.6 & 81.9 & 7.6 & 0.792 & 0.819 & 1.4 & 3.3 \\
LL-SALUN0.1 + LipLoss($n=25$, $std=0.5$) $lr=1\times10^{-4}$ $\lambda_{\text{retain}}=0.0$ 5 Iters & 47.3 & 89.9 & 17.5 & 0.749 & 0.762 & 10.1 & 17.2 \\
EN-SALUN0.5 + LipLoss($n=25$, $std=0.1$) $lr=1\times10^{-4}$ $\lambda_{\text{retain}}=0.0$ 1000 Iters & 6.9 & 33.3 & 1.5 & 0.382 & 0.328 & 0.0 & 0.2 \\
EN-SALUN0.5 + LipLoss($n=25$, $std=0.5$) $lr=1\times10^{-4}$ $\lambda_{\text{retain}}=0.0$ 50 Iters & 38.2 & 82.6 & 7.7 & 0.787 & 0.814 & 0.7 & 3.8 \\
EN-SALUN0.1 + LipLoss($n=25$, $std=0.5$) $lr=1\times10^{-4}$ $\lambda_{\text{retain}}=0.0$ 50 Iters & 38.1 & 82.6 & 7.8 & 0.788 & 0.815 & 0.7 & 3.7 \\
EN-SALUN0.5 + LipLoss($n=25$, $std=0.1$) $lr=1\times10^{-4}$ $\lambda_{\text{retain}}=1.0$ 1000 Iters & 83.9 & 97.1 & 97.2 & 0.600 & 0.643 & 89.1 & 96.4 \\
EN-SALUN0.5 + LipLoss($n=25$, $std=0.1$) $lr=1\times10^{-4}$ $\lambda_{\text{retain}}=0.5$ 1000 Iters & 83.0 & 97.1 & 97.4 & 0.601 & 0.647 & 76.1 & 96.4 \\
EN-SALUN0.5 + LipLoss($n=25$, $std=0.1$) $lr=1\times10^{-4}$ $\lambda_{\text{retain}}=0.1$ 1000 Iters & 82.3 & 97.1 & 97.1 & 0.603 & 0.650 & 52.9 & 96.3 \\
\rowcolor{gray!12}
EN-SALUN0.5 + LipLoss($n=25$, $std=0.1$) $lr=1\times10^{-4}$ $\lambda_{\text{retain}}=0.05$ 1000 Iters & 81.5 & 96.4 & 96.7 & 0.601 & 0.655 & 44.2 & 96.2 \\
\hline
\rowcolor{gray!12}
RL $lr=1\times10^{-4}$ $\lambda_{\text{retain}}=1.0$ 25 Iters & 87.3 & 97.1 & 97.3 & 0.636 & 0.624 & 0.0 & 96.5 \\
RL $lr=1\times10^{-4}$ $\lambda_{\text{retain}}=1.0$ 250 Iters & 87.7 & 97.1 & 97.7 & 0.783 & 0.628 & 2.2 & 96.5 \\
RL $lr=1\times10^{-4}$ $\lambda_{\text{retain}}=1.0$ 500 Iters & 87.9 & 97.1 & 97.6 & 0.839 & 0.635 & 2.2 & 96.6 \\
RL $lr=1\times10^{-4}$ $\lambda_{\text{retain}}=0.5$ 250 Iters & 87.7 & 97.1 & 97.5 & 0.812 & 0.636 & 0.7 & 96.5 \\
RL $lr=1\times10^{-4}$ $\lambda_{\text{retain}}=0.1$ 250 Iters & 87.9 & 97.8 & 94.4 & 0.883 & 0.723 & 0.0 & 96.4 \\
RL $lr=1\times10^{-4}$ $\lambda_{\text{retain}}=0.0$ 250 Iters & 64.7 & 96.4 & 15.4 & 0.942 & 0.919 & 0.0 & 15.1 \\
RL $lr=1\times10^{-4}$ $\lambda_{\text{retain}}=0.0$ 25 Iters & 83.9 & 96.4 & 91.0 & 0.675 & 0.632 & 0.0 & 85.5 \\
RL $lr=1\times10^{-4}$ $\lambda_{\text{retain}}=0.1$ HeadFreeze 250 Iters & 88.1 & 97.8 & 97.0 & 0.846 & 0.658 & 0.0 & 96.4 \\
\hline
GA $lr=1\times10^{-4}$ $\lambda_{\text{retain}}=0.0$ 10 Iters & 86.8 & 97.1 & 97.1 & 0.665 & 0.665 & 0.0 & 96.2 \\
GA $lr=1\times10^{-4}$ $\lambda_{\text{retain}}=0.0$ 25 Iters & 88.1 & 97.8 & 87.8 & 0.927 & 0.788 & 0.0 & 39.6 \\
GA $lr=1\times10^{-5}$ $\lambda_{\text{retain}}=0.0$ 25 Iters & 85.9 & 97.1 & 97.5 & 0.531 & 0.637 & 59.4 & 96.5 \\
\rowcolor{gray!12}
GA $lr=1\times10^{-5}$ $\lambda_{\text{retain}}=0.0$ 50 Iters & 87.3 & 97.1 & 97.2 & 0.681 & 0.663 & 0.0 & 96.4 \\
\hline
\rowcolor{gray!12}
Contrastive($\tau=0.1$) $lr=1\times10^{-4}$ $\lambda_{\text{retain}}=0.0$ 250 Iters & 87.7 & 97.1 & 97.5 & 0.603 & 0.619 & 97.1 & 96.6 \\
Contrastive($\tau=0.1$) $lr=1\times10^{-3}$ $\lambda_{\text{retain}}=0.0$ 250 Iters & 2.7 & 16.7 & 0.7 & 0.998 & 0.999 & 0.0 & 0.0 \\
Contrastive($\tau=0.1$) $lr=1\times10^{-3}$ $\lambda_{\text{retain}}=0.0$ 100 Iters & 23.7 & 66.7 & 1.6 & 0.663 & 0.591 & 10.1 & 4.3 \\
Contrastive($\tau=0.1$) $lr=1\times10^{-3}$ $\lambda_{\text{retain}}=0.0$ 50 Iters & 87.7 & 97.1 & 97.6 & 0.605 & 0.618 & 97.1 & 96.6 \\
Contrastive($\tau=0.1$) $lr=1\times10^{-3}$ $\lambda_{\text{retain}}=0.1$ 100 Iters & 88.0 & 97.1 & 97.6 & 0.625 & 0.623 & 97.1 & 96.6 \\
Contrastive($\tau=0.1$) $lr=1\times10^{-3}$ $\lambda_{\text{retain}}=0.05$ 100 Iters & 87.5 & 97.1 & 97.5 & 0.640 & 0.614 & 96.4 & 96.5 \\
Contrastive($\tau=0.1$) $lr=1\times10^{-3}$ $\lambda_{\text{retain}}=0.001$ 100 Iters & 17.6 & 59.4 & 1.3 & 0.774 & 0.867 & 1.4 & 0.2 \\
Contrastive($\tau=0.2$) $lr=1\times10^{-3}$ $\lambda_{\text{retain}}=0.0$ 100 Iters & 87.8 & 97.1 & 97.5 & 0.608 & 0.616 & 97.1 & 96.6 \\
Contrastive($\tau=0.5$) $lr=1\times10^{-3}$ $\lambda_{\text{retain}}=0.0$ 100 Iters & 87.7 & 97.1 & 97.5 & 0.602 & 0.619 & 97.1 & 96.6 \\
\hline
\end{tabular}
}
\caption{Results for a lightweight hyperparameter search for each baseline. Rows highlighted in gray correspond to the configurations reported in the main paper.}
\label{tab:celeba_hyp_search}
\end{table*}

\end{document}